\documentclass[runningheads]{llncs}
\usepackage[T1]{fontenc}
\usepackage{graphicx}
\usepackage{multirow}
\usepackage{adjustbox}
\usepackage{xcolor}
\usepackage{colortbl}
\newcommand{\red}[1]{\textcolor{red}{#1}}

\newcommand{\blue}[1]{\textcolor{blue}{#1}}
\newcommand{\onedot}{.}
\newcommand{\revised}[1]{\textcolor{black}{#1}}
\usepackage{paralist}
\usepackage{booktabs}
\usepackage{arydshln}
\usepackage{adjustbox}
\usepackage{subcaption}
\usepackage{tabularx} 
\usepackage{diagbox}
\usepackage{amsmath}
\usepackage{amssymb}
\usepackage{wrapfig}
\usepackage{algorithm}
\usepackage{algpseudocode}
\usepackage{rotating} 
\usepackage{booktabs} 
\usepackage{pifont}
\usepackage[colorlinks=true, linkcolor=blue, citecolor=blue, urlcolor=blue]{hyperref}

\def\ie{\emph{i.e}\onedot}

\def\etal{\emph{et al}\onedot}
\makeatletter
\renewcommand*{\@fnsymbol}[1]{\ensuremath{\ifcase#1\or *\or \dagger\or \ddagger\or
    \mathsection\or \mathparagraph\or \|\or **\or \dagger\dagger
    \or \ddagger\ddagger \else\@ctrerr\fi}}
\makeatother

\begin{document}
\title{LF Tracy: A Unified Single-Pipeline Paradigm for Salient Object Detection in Light Field Cameras}

\titlerunning{LF Tracy}

\author{Fei Teng\inst{1,}\thanks{Equal contribution.} \and
Jiaming Zhang\inst{2,*} \and
Jiawei Liu\inst{1} \and 
Kunyu Peng\inst{2} \and
Xina Cheng\inst{3} \and
Zhiyong Li\inst{1} \and Kailun Yang\inst{1,}\thanks{Corresponding author: kailun.yang@hnu.edu.cn}}

\authorrunning{F. Teng et al.}

\institute{Hunan University, Changsha, China \and Karlsruhe Institute of Technology, Karlsruhe, Germany \and Xidian University, Xi'an, China
}
\maketitle              

\begin{abstract}
Leveraging rich information is crucial for dense prediction tasks. Light field \text{(LF)} cameras are instrumental in this regard, as they allow data to be sampled from various perspectives. This capability provides valuable spatial, depth, and angular information, enhancing scene-parsing tasks. However, we have identified two overlooked issues for the LF salient object detection \text{(SOD)} task. \text{(1):} Previous approaches predominantly employ a customized two-stream design to discover the spatial and depth features within light field images. The network struggles to learn the implicit angular information between different images due to a lack of intra-network data connectivity. \text{(2):} Little research has been directed towards the data augmentation strategy for LF SOD. Research on inter-network data connectivity is scant. In this study, we propose an efficient paradigm (LF Tracy) to address those issues. This comprises a single-pipeline encoder paired with a highly efficient information aggregation (IA) module \text{(${\sim}8M$ parameters)} to establish an intra-network connection. Then, a simple yet effective data augmentation strategy called MixLD is designed to bridge the inter-network connections. Owing to this innovative paradigm, our model surpasses the existing state-of-the-art method through extensive experiments. Especially, LF Tracy demonstrates a $23\%$ improvement over previous results on the latest large-scale PKU dataset. The source code is publicly available at:~\url{https://github.com/FeiBryantkit/LF-Tracy}.

\keywords{Light field camera  \and salient object detection \and neural network \and scene parsing.}
\end{abstract}

\section{Introduction}

The objective of SOD lies in mimicking human visual attention mechanisms to accurately identify the most conspicuous objects or regions in a variety of visual contexts. In particular, SOD plays a dual role: it not only aids agents in discerning the most striking and important elements in visual scenarios but also plays a pivotal role in several downstream tasks, including object detection, segmentation, and other dense prediction tasks~\cite{wang2015saliency,borji2012state}. 

\begin{wrapfigure}{l}{0.48\textwidth}
\vspace{-2.2em}
  \centering
  \includegraphics[width=0.46\textwidth]{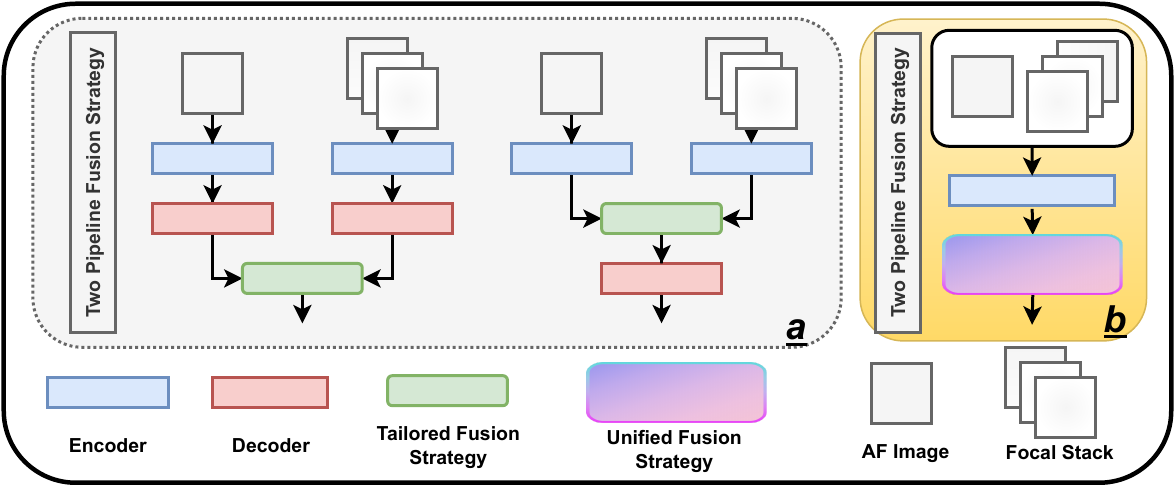}
  \caption{Paradigms of LFSOD model. The conventional two-stream methods ($\underline{a}$) and our single-pipeline method ($\underline{b}$).}
  \label{fig:depth_light}
  \vspace*{-2em}
\end{wrapfigure}

Within the SOD community, the current 2D-based methods~\cite{pang2020multi,chen2020global} rely on the powerful feature extraction capabilities of Convolutional Neural Networks and Transformers, coupled with finely crafted decoders, to achieve impressive results. Meanwhile, a rich array of 3D methods~\cite{sun2023catnet,chen2022modality} have been introduced by utilizing depth or thermal information to boost the result. Given that information from various domains aids neural networks in more effectively learning scene features, LF cameras have been introduced~\cite{georgiev2006light}. LF camera is capable of capturing spatial, depth, and angular information. However, two significant challenges are neglected.

\textit{One: Lacking Intra-network Data Connectivity.}
\revised{The existing datasets for LF cameras consist of post-processed All-Focused (AF) images and Focal Stacks (FS) \cite{LFSD,duft,huft,gao2023thorough}. AF images are full of texture information. FS images refer to images that include angular and depth information.}
The asymmetric data construction enriches the geometric information captured by LF cameras. 

\begin{wrapfigure}{l}{0.35\textwidth}
\vspace{-2.2em}
  \centering
  \includegraphics[width=0.34\textwidth]{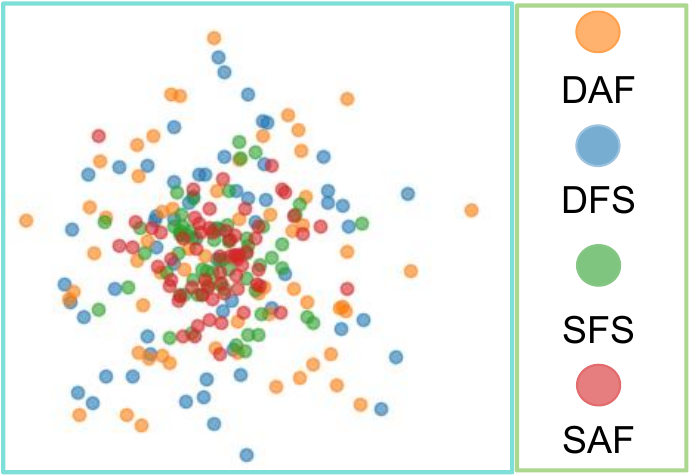}
  \caption{Search space is visualized utilizing TSNE. ``DAF'' and ``DFS'' represent the feature maps of AF and FS in the dual pipeline method, while ``SAF'' and ``SFS'' represent the feature maps of AF and FS in the single pipeline method.}
  \label{fig:Search Space}
  \vspace*{-2.2em}
\end{wrapfigure}

\revised{However, the implicit angular details cannot be directly utilized; they can only be obtained by exploring the latent relationships between images. While effectively utilizing depth and spatial information enhances the network's ability to understand scenes, the current two-stream approach (Fig.~\ref{fig:depth_light}(a)) neglects essential linkages among various images and disregards the angular information flow throughout the network, resulting in smaller searching space. As illustrated in Fig.~\ref{fig:Search Space}, the high-dimensional data visualization (\ie, TSNE) is conducted to demonstrate the search space of features. The search space of SFS (single-pipeline, focal stack) and SAF (single-pipeline, all-focused image) is significantly larger than that of the two-stream method.}

Furthermore, in using a single-pipeline encoder, while different images can guide the network to learn angular features, merging the unfocused segments in AF image with all-focused data in FS images results in feature contamination within the feature space, significantly undermining the network's discriminative capabilities. Hence, one of the key points of our work is \textbf{``how to leverage angular information while circumventing the alignment issues brought about by varying shooting viewpoints?''}.

\begin{wrapfigure}{r}{0.48\textwidth}
  \centering
  \includegraphics[width=0.46\textwidth]{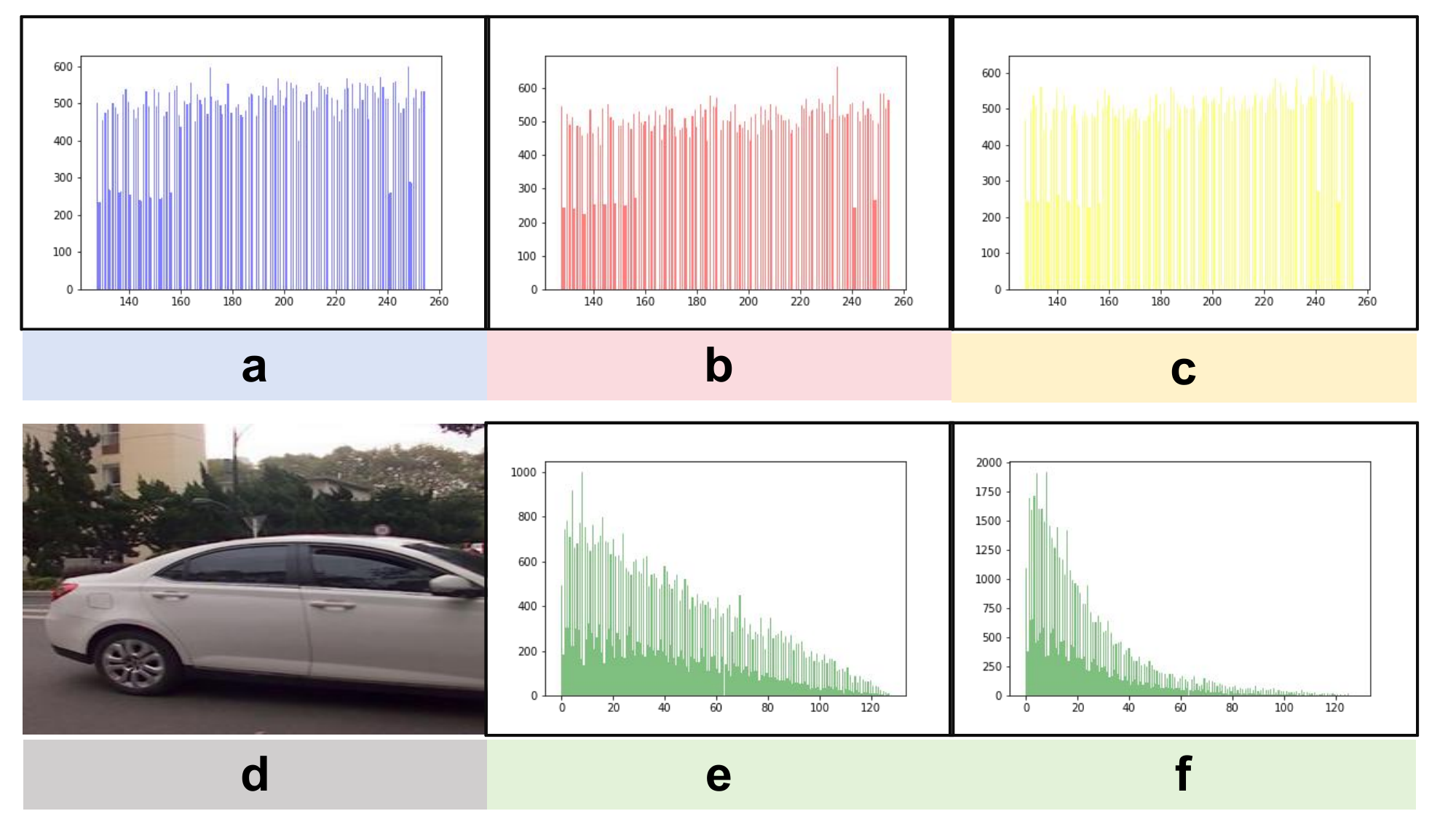}
  \caption{A statistical result for the MixLD strategy. (d) indicates the AF image, and (a) and (b) illustrate the pixel distribution of the AF image and the FS image, respectively. (c) represents the central image after MixLD. (e) and (f) show the difference maps between the AF information and the original FS image, before and after MixLD. In (a), (b), and (c), the horizontal and vertical axes represent the pixel values and the number of pixels at those values, respectively. For (e) and (f), they represent the pixel value differences and the number of pixels at those difference values, respectively. }
  \label{fig:DA}
  \vspace*{-2em}
\end{wrapfigure}
\textit{Two:} \textit{Lacking Inter-network Data Connectivity.}
Although researchers in the LF community enhance the understanding of scenes by introducing depth information (Focal Stack), existing works still adhere to the conventional RGB-D fusion structures~\cite{10168184,10122696}, employing common data augmentation \text{(DA)} strategies. Those methods isolatedly excavate the angular features and bury the relationship between different LF representations since there is no data interaction before the training process~\cite{xiao2023cutmib}. Therefore, another key point of our work is \textbf{``developing a novel DA strategy specifically for the LFSOD task to bridge a connection between various LF data sources before the training process''}. Fig.~\ref{fig:DA} indicates a statistical result through the MixLD strategy. Before applying data augmentation, although a certain degree of data similarity between AF and FS can be observed from figures Fig.~\ref{fig:DA}(a) and Fig.~\ref{fig:DA}(b), there are still considerable differences in data within the range of $[0,100]$ pixels. However, after DA, by analyzing the distribution of the phase spectrum (Fig.~\ref{fig:DA}(c)) and calculating its similarity with the central figure in the frequency domain (Fig.~\ref{fig:DA}(f)), it can be seen that information has been aggregated.

In this work, we propose a novel paradigm (LF Tracy) to overcome the aforementioned challenges. Firstly, a single-pipeline framework in Fig.~\ref{fig:depth_light}(b) is established to achieve the intra-network data connectivity. By learning different LF representations from a comprehensive perspective through a single backbone, our network can fully utilize the information from LF images rather than conducting separate feature extraction for LF representations. Furthermore, a simple yet IA model is performed within LF Tracy to effectively align and fuse the coupled features through the same backbone. Moreover, a simple data augmentation strategy called MixLD is introduced to establish inter-network data connectivity. 

To demonstrate the efficiency of the proposed LF Tracy paradigm, comprehensive experiments are conducted on the large-scale PKU dataset~\cite{gao2023thorough}, which comprises samples from both terrestrial and aquatic environments, and the LFSOD datasets~\cite{LFSD,duft,huft}.  
By employing this paradigm ($\text{MixLD} + \text{Backbone} + \text{IA}$), our network achieved the state-of-the-art performance compared with previous works. Specifically, on the PKU dataset, our work achieved a $23\%$ improvement in accuracy, fully validating the effectiveness of our network.

At a glance, we deliver the following contributions:
\begin{compactitem}
    \item \revised{We propose a single-stream SOD paradigm from scratch, bridging the inter-network and intra-network data connectivity.}
    \item We have designed a low-parameter Information Aggregation (IA) Module that uncovers angular information while avoiding feature aliasing. Furthermore, we introduce a data augmentation strategy, namely MixLD, to establish inter-network data connectivity.
    \item \revised{An in-depth analysis is conducted to evaluate the performance of the single stream network under different hyper-parameters and module combinations.}
    \item Our method achieves top performance on three LF datasets and one large-scale PKU dataset, which comprises over $10K$ images.

\end{compactitem}
 \

%%%%%%%%%%%%%%%%%%%328
\vspace{-1ex}
\section{Related Work}
\vspace{-1ex}
Discovering and connecting the spatial, depth, and angular information of LF is essential for designing an efficient SOD neural network. Therefore, we will discuss the utilization of light field information from two aspects: \textbf{Intra-network Data Connectivity} in Sec.~\ref{2.1} and \textbf{Inter-network Data Connectivity} in Sec.~\ref{2.2}. Lastly, preliminaries related to LF imaging are introduced in the appendix.

\vspace{-1em}
\subsection{Intra-network Data Connectivity} \label{2.1}

The SOD task can be traced back to rule-based methodologies, which predominantly relied on visual attributes such as color, contrast, and spatial distribution to ascertain salient areas in images.%~\cite{itti1998model,achanta2008salient}. 
~In recent years, there has been a paradigm shift in the SOD community towards leveraging deep learning paradigms. 
Specifically, MENet~\cite{wang2023pixels} introduced iterative refinement and frequency decomposition mechanisms to improve detection accuracy. By utilizing transformer and multi-scale refinement architecture, Wang~\etal~\cite{deng2023recurrent} used high- and low-resolution images to achieve SOD. Furthermore, Zhang~\etal~\cite{cong2023multi} implemented SOD for panoramic images. Apart from those single-modality SOD networks, depth information is introduced to enhance performance, whereas multi-model fusion strategies~\cite{chen2022modality,cong2022does} are employed for RGB and thermal data. 

For the SOD task of LF, Wang~\etal~\cite{DLLFSD} implemented a dual-pipeline neural network in the SOD community. Since then, the two-stream approach~\cite{jing2021occlusion} for processing LF images has stood in a leading position in this field. 
Typically, this involves employing one backbone for processing AF images and another for FS images or the depth image extracted from LF sub-aperture images. Although the two-stream approach has seen considerable advancement in various tasks~\cite {zhang2023delivering}, it is typically applied to modalities that are isolated, such as depth and RGB images. For light field cameras, the depth, angular, and spatial information are embedded across different representations,~\ie,~AF images and FS images. Processing these images in an isolated manner buries the angular features of light field cameras, and thus remains a sub-optimal method~\cite{xiao2023cutmib}. 

\vspace{-1em}
\subsection{Inter-network Data Connectivity} \label{2.2}

Data augmentation (DA) has been thoroughly explored in various vision tasks such as image recognition, image classification, and semantic segmentation, proving effective in enhancing network performance and mitigating the issue of overfitting.~%~\cite{upchurch2017deep,xiao2023cutmib}. 
The traditional data augmentation strategies can be roughly divided into five categories based on the adjusted purpose. 
1) Flipping the image along its vertical and horizontal axis is a typical technique for increasing the diversity of data available for training. Furthermore, rotating an image at a certain angle is also a contributing factor.
2) Color jitter simulates images under different lighting and camera settings, enabling the trained model to better adapt to various scenarios. 
3) Cutout~\cite{devries2017improved} is introduced to drought or mismatch part of pixel-level information between neighboring pixels to increase the discrimination capability of the network. 
4) Beyond deep learning, several works~\cite{Reinforce1,zhang2023bo} introduced machine learning-based strategies to boost the network capability. 
5) Mixing-based methods~\cite{xiao2023cutmib,florea2023softclustermix} leverage information from multiple images by generating blended input images. 

Those methods demonstrate noticeable performance for the single image in the augmentation community. However, for light field cameras, the subtle angular information hidden within the interplay of multiple images cannot be captured through DA applied to individual images alone. Thus, establishing data connectivity across networks becomes crucial. 
 \
\vspace{-2em}
\section{Methodology}

This section introduces a comprehensive overview of our proposed paradigm, designed for the LFSOD task. 
Firstly, the framework's architecture is meticulously expounded in Sec.~\ref{3.1}. Additionally, in Sec.~\ref{3.2}, we introduce a simple yet fusion module, which is pivotal for efficiently aggregating Light Field features. Last but not least, Sec.~\ref{3.3} delves into our innovative DA Strategy.

\vspace{-1ex}
\begin{figure}[b]
  \centering
  \includegraphics[width=0.98\textwidth]{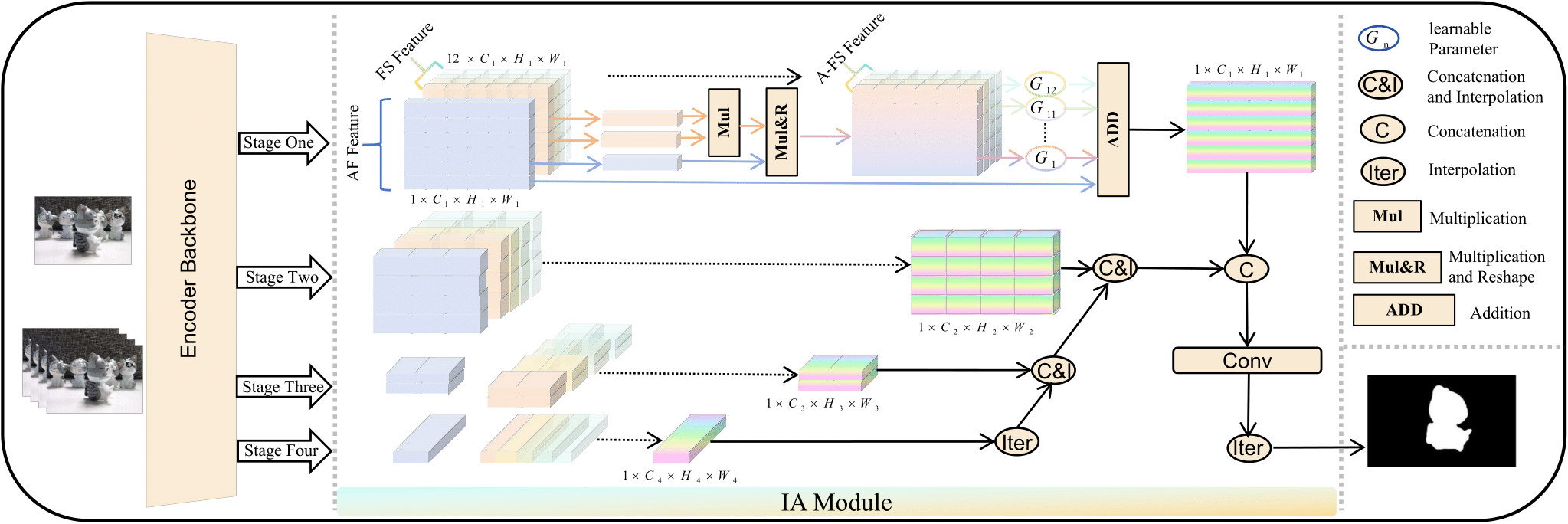}
  %\vspace*{1mm}
  \caption{Pipeline of LF Tracy network. FS and AS images are fed into the backbone for feature extraction. Multi-level features are then fed into the IA module (Sec.~\ref{3.3}) for in-network data fusion and to predict the final result image.}
  \label{fig:Overall}
  \vspace{-4em}
\end{figure}

\vspace{-1ex}
\subsection{Proposed LF Tracy Framework} \label{3.1}
As shown in Fig.~\ref{fig:Overall}, the proposed network has two components: a four-stage encoder providing rich multi-dimensional information from different asymmetric data and the IA Module. 
The IA Module serves a dual purpose: 
1) It overcomes the mismatching between the features established in-network connectivity through the same encoder block. 
2) It can realign these features before sending them to the prediction head. The AFttention image $I_{AF}^m$ and FSttention stack $I_{FS}^m$ are described separately to provide a more intuitive description of the information flow and interaction process. The AFttention image and FSttention stack indicate the data source after MixLD. Furthermore, for simplicity, the following description is based on the stage one, which is the same for the other three stages. Especially, by applying the encoder block, the images are transferred into AF features $({F_{AF}}{\in}\mathbb{R}^{(64 {\times} 64 {\times} 64)})$ and FS features $({F_{FS}^n}{\in}\mathbb{R}^{(64 {\times} 64 {\times} 64)}|{n\in [1,12]})$.
After applying the IA Module, the $13$ features are aggregated into one feature $(f_{1}{\in}\mathbb{R}^{(64 {\times} 64 {\times} 64)})$, which contains all spatial, angular, and depth information. After four stages, there is a set of feature maps $\{f_l|l \in {[1,4]}\}$ with channel dimension $\{64,128,320,512\}$. Only $f_1$ is described in detail here, as the processes for the other dimensions are identical. Furthermore, at the training stage, to cooperate with the structure loss~\cite{fan2017structure} calculation, $f_{l}$ is also passed through one convolutional layer to compress channel information, as in Eq.~(\ref{CC}).
\begin{align}
    f_{M_1} = Conv(64,1)(f_{l}),\label{CC}
\end{align}
where $Conv(64,1)(\cdot)$ indicates the convolutional layer with input channel $64$ and output channel one. $f_{M_1}$ denotes the feature after merging at the first stage. Furthermore, drawing upon the structure loss as outlined in~\cite{wei2020f3net}, we have integrated the Tversky Loss~\cite{salehi2017tversky} into our training process to improve supervision during training, specifically targeting a reduction in false positives and negatives. 

\vspace{-1em}
\subsection{Information Aggregation: IA Module} \label{3.2}
To fuse the implicit angular, explicit spatial, and depth information from asymmetric data, we introduce a simple IA Module that follows a two-step interaction process. Firstly, given single feature $(F_{FS}^n| n \in [1,12])$, the FS-guided Querry and Key are generated through their respective convolutional layer. Through matrix multiplication, the attention map $(M{\in}\mathbb{R}^{(4096 {\times} 4096)})$ is obtained. The attention map integrates a broader context into the aggregation of local features and enhances the representative capability of the focus part. Furthermore, applying the third convolutional layer to $F_{AF}$, the AF image guided Value $(V_{AF}{\in}\mathbb{R}^{4096 {\times} 64})$ is generated, as in~Eq.~(\ref{1})-(\ref{4}). \revised{Given that the SOD task is sensitive to hyper-parameters and module design, we adopt a dimension reduction method for Query and Key. For more details on dimension reduction, please refer to Sec.~\ref{AB:3}.}
\begin{align}
    Q = Conv_{q}(C_{in},C_{out}^*)(F_{FS}^n),\label{1}  \\
    K = Conv_{k}(C_{in},C_{out}^*)(F_{FS}^n),\label{2} \\
    M = Soft\{Mul(Q,K)\},\label{3} \\
    V = Conv_{v}(C_{in},C_{out})(F_{AF}). \label{4}
    \vspace{-1em}
\end{align}
The tokens $(T{\in}\mathbb{R}^{4096 {\times} 64})$, which contains information from certain focal images, is obtained by multiplication of attention map and Value, as in Eq.~(\ref{Fusion3}).
\begin{align}
     T = Mul(M,V). \ \label{Fusion3}
\end{align}
After obtaining the tokens, the A-FS features $\hat{F}_{FS}^n$ are generated by applying the reshape operation. Note that the number of images has remained unchanged until now. This operation aims to enhance the spatial information at the corresponding depth by guiding the information from the FS image and, with the help of AF features, establish a connection between the global AF and FS information. Secondly, we introduce a set of leanable parameters to calculate the contribution of different FS features. To further enhance the spatial context information, a submission is undertaken, and the final result $f_{1}$ is obtained as in Eq.~({\ref{5}}).
\begin{align}
    f_{1} = {AF} + \sum_{i=1}^{12}\sigma \times \hat{F}_{FS}^n. \label{5}
\end{align}

Given multi-scale features $\{f_{1}, f_{2}, f_{3}, f_{4}\}$, interpolation and concatenation are conducted. Finally, by applying the convolutional layer following an interpolation, the mask $f$ is compressed and sent to the prediction head.

\vspace{-0.5em}
\subsection{Data Augmentation Strategy: MixLD}~\label{3.3}
\begin{wrapfigure}{r}{0.42\textwidth}
\vspace{-1.8em}
  \centering
  \includegraphics[width=0.42\textwidth]{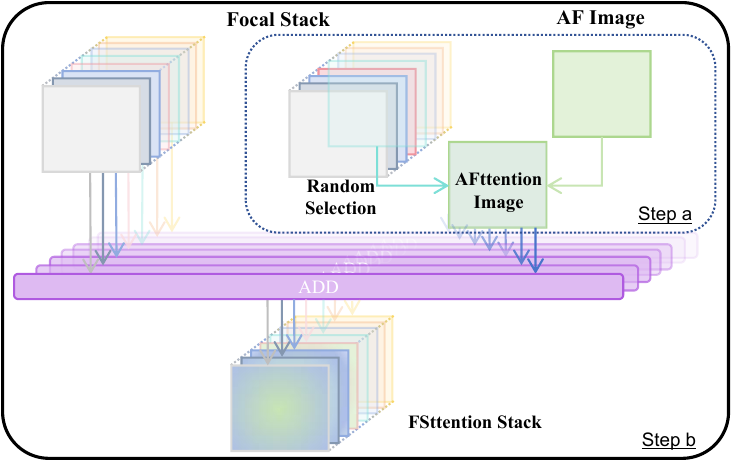}
  \caption{Schematic illustration of our proposed MixLD strategy tailored for LFSOD. The strategy contains two independent steps ($\underline{a}$ and $\underline{b}$), each of which is carried out randomly.}
  \label{fig:MixUp}
  \vspace{-2em}
\end{wrapfigure}

As depicted in Fig.~\ref{fig:MixUp}, the primary objective of the specific data augmentation strategy for the LFSOD task is to amalgamate distinct representations inherent in light field camera, namely, AF image $(I_{AF})$, FS~$(I_{FS}^n | n \in [1,12])$, and implicit angular information. This strategy is methodically partitioned into two discrete phases, each targeting specific aspects of the integration process. Initially, a non-intrusive approach is employed to integrate angular and depth information into the composite AF image while preserving the integrity of spatial data dimensions. 
Specifically, the data augmentation strategy can be described as following steps:

\textbf{Firstly: \revised{(FS2AF)}} Following the FS setting~\cite{piao2021panet}, one FS slice $I_{FS}^n$ with dimension $\{3{\times}256{\times}256\}$ is randomly selected with a likelihood of $0.1$. This FS image is then subjected to a pixel-level fusion process, meticulously blending it into the AF image representation, as shown in Eq.~(\ref{mix1}).
\begin{align}
I_{AF}^m &= \{\alpha \times {I_{AF} + (1-\alpha)\{ \text{Rand}(I_{FS}^n}) \},\label{mix1} 
\end{align}
where $\alpha$ denotes the degree of blending and $n$ indicates the quantities of focal images. $I_{AF}^m$ indicates the AF image after blending, \ie, AFttention image. In MixLD, $\alpha{=}1$ indicates no blending and $\alpha{=}0$ indicates that the AF Image is completely replaced. Only the AF image is altered during this process, while the FS images remain unchanged. Meanwhile, this procedure is not conducted for each interaction. 

\textbf{Secondly: \revised{(AF2FS)}} The AFttention image is integrated into all the FS images with a probability of $0.5$, as in Eq.~(\ref{mix2}).
\begin{align}
    I_{F_n}^m = \{\beta \times {I_{AF}^m} + (1-\beta) \times I_{FS}^n\},\label{mix2} 
\end{align}
where $\beta$ denotes also a super parameter for the degree of blending in stage two and $n$ denotes the quantities of focal images. $I_{F_n}^m$ indicates the FS after blending \ie, FSttention stack. This integration carried out with a fusion probability of $0.5$ instead of $0.1$, aims to make it more possible to enrich the FS with additional information. By blending the AF image into the FS images, each focal image retains its inherent depth information, gains implicit angular insights from the other focal image, and enhances its spatial geometric information from the AF image. Furthermore, the AFttention image $I_{AF}^m$ and FSttention stack $(I_{F_n}^m| n \in [1,12])$ are fed into the network. \revised{It is important to emphasize that both phases (FS2AF and AF2FS) of MixLD are conducted randomly. It is possible for data interaction to occur in only one phase, while the other remains non-interactive.} 

It is precisely through this form of blending that the neural network while learning the inherent AF and FS information, can break out of the conventional framework to learn implicit angular information. For detailed algorithms, please refer to the pseudocode presented in the Appendix.
\vspace{-1em}

 \

\vspace{-1em}
\section{Experiments}

To effectively demonstrate the efficacy of the approach, we showcase the quantitative result and qualitative results on different datasets. Firstly, we introduce the experimental setup in Sec.~\ref{4.1}. Secondly, in Sec.~\ref{4.2}, we present a quantitative comparison with other methods. Thirdly, in Sec.~\ref{4.3}, we showcase the visual results of the method, along with a visual comparison with previous approaches.

\subsection{Implementation Details} \label{4.1} 

\textit{Datasets:} \revised{The experiments are conducted following the benchmark proposed by the PKU team~\cite{gao2023thorough}. The datasets involve traditional LFSOD datasets, which include LFSD~\cite{LFSD}, DUT-LF~\cite{duft}, HFUT~\cite{huft} and a large-scale PKU dataset~\cite{gao2023thorough}.} The images within the PKU dataset are sourced from terrestrial and aquatic environments. Two experiment strategies are conducted: \textbf{I)} training on DUT-LF {$+$} HFUT, ${\sim}1000$ images, and evaluation on the whole LFSD dataset, the DUT-LF testing dataset, and HFUT testing dataset; \textbf{II)} training and testing on the PKU-LF dataset. PKU-LF dataset contains more than $10K$ images. For the ablation study, the experiments are based on experiment strategy one.

\textit{Setting Details:} The image size for all the datasets is $256{\times}256$. Each scene is structured to contain exactly $12$ focal slices to meet specific coding requirements. This is achieved by strategically duplicating focal slices in the original order. Data augmentation is applied with Flipping, Cropping, Rotating, and MixLD for the training process. The blending parameter $\alpha, \beta$ are set into $0.5$ and $0.5$, respectively. The AdamW optimizer with a learning rate of $5e^{-5}$ and weight decay of $1e^{-4}$ is adapted for training. All the experiments are conducted on one A6000 GPU with a batch size of $6$. The training epochs are limited to $300$.

\textit{Evaluation Metrics:} To analyze the results of different methods, we employ mean absolute error (MAE)~\cite{perazzi2012saliency} for a fair comparison. \revised{For F-measure ($F_\beta^{mean}$)~\cite{achanta2009frequency}, E-measure ($S_\beta^{man}$)~\cite{fan2017structure}, S-measure ($S_{\alpha}$), we compare them with the previously best methods.}

\begin{table}[t]
 \centering
  \LARGE
  \renewcommand{\arraystretch}{1.2}
  \begin{adjustbox}{width=0.98\textwidth}
\begin{tabular}{l:ccccccccc:ccc:cc}

\toprule[2mm]
\multirow{2}{*}{Methods}&LFNet& SSF & D3Net & ATSA & UCNet& ESCNet & JLDCF &MEANet& GFRNet   & \multicolumn{1}{:c}{STSA$_{1}$} & STSA$_{2}$ & STSA$_{3}$& \textbf{Ours} &\textbf{Gain}\\ 
\vspace{-2mm}
&\cite{zhang2020lfnet}&\cite{zhang2020select}&\cite{fan2020rethinking}&\cite{zhang2020asymmetric}&\cite{zhang2021uncertainty}&\cite{zhang2022exploring}&\cite{fu2021siamese}&\cite{jiang2022meanet}&\cite{yuan2023guided}&\multicolumn{3}{:c:}{\cite{gao2023thorough}}& & \\
&~TIP20&~CVPR20&~TNNLS21&~ECCV21&TPAMI22&TIP22&TPAMI22&Neuc22&ICME23&\multicolumn{3}{:c:}{TPAMI23}& & \\

\midrule \midrule
LFSD         & .092 & .067 & .095  & .068 & .072  & n.a.  & .070 & .077 & .065 & .067    & .065    & \blue{.062} &  \red{.046} & \red{\textbf{26\%$\uparrow$}}\\

HFUT        & .096 & .100 & .091  & .084 & .090  & .090  & .075 & .072 & .072  & .067    & .072    & \blue{.057}  & \red{.056} &\red{\textbf{2\%$\uparrow$}}\\
DUT-LF     & .055 & .050 & .083  & .041 & .081   & .061 & .058  & .031& \blue{.026}  & .033    & .030    & .027  & \red{.023} &\red{\textbf{12\%$\uparrow$}}\\
PKU-LF    & {n.a.} & .062 & .067  & .045 & .070  & n.a. & .049 & {n.a.}  & {n.a.} & .047    & .042    & \blue{.035}  & \red{.027}&\red{\textbf{23\%$\uparrow$}}\\ \hline

\end{tabular}
\end{adjustbox}
\vspace*{2mm}
\caption{Quantitative comparison with other methods in terms of MAE. The best result is highlighted in red. ``Gain'' indicates the improvement in our results compared to previous state-of-the-art methods. STSA$_1$, STSA$_2$, and STSA$_3$ represent the outcomes of the PKU Team~\cite{gao2023thorough} using different quantities of data for training. Although STSA$_3$ uses DUFT+HFUT+PKU-LF as a training set and outperforms other methods, our method still surpasses the STSA$_3$ network without expanding Training data. ``Gain'' denotes unavailable results.}
\label{tab:propotion}
\vspace{-1em}
\end{table}

\vspace{-1em}
\subsection{Quantitative Results} \label{4.2}
\vspace{-1em}

To verify the efficiency of the approach, we compare the designed network with existing methods. 
Table~\ref{tab:propotion} shows that the best performance of the proposed approach significantly outperforms existing methods across the LFSD series dataset~\cite{LFSD,duft,huft} and PKU dataset~\cite{gao2023thorough} on MAE. Due to the variability in performance across different evaluation metrics and datasets, we follow the benchmark provided by the PKU team~\cite{gao2023thorough}. 

\begin{wraptable}{r}{0.55\textwidth}
\vspace{-2em}
  \centering
  \scriptsize
  \setlength{\tabcolsep}{1mm}
  \begin{tabular}{cc|ccc}
    \toprule[0.5mm]
    Dataset & Metrics & PreV & Our & Gain \\
    \midrule \midrule
    \multirow{3}{*}{LFSD~\cite{LFSD}} & $F_\beta^{mean}$ & .862~\cite{10122696} & .896 & \red{\textbf{3.9\%$\uparrow$}} \\
                          & $E_\beta^{mean}$ & .902~\cite{gao2023thorough} & .912 & \red{\textbf{1.1\%$\uparrow$}} \\
                          & $S_{\alpha}$ & .864~\cite{fan2020bbs} & .902 & \red{\textbf{4.4\%$\uparrow$}} \\
    \midrule
    \multirow{3}{*}{HFUT~\cite{huft}} & $F_\beta^{mean}$ & .771~\cite{gao2023thorough} & .769 & \blue{\textbf{0.3\%$\downarrow$}}\\
                          & $E_\beta^{mean}$ & .864~\cite{gao2023thorough} & .865 & \red{\textbf{0.1\%$\uparrow$}} \\
                          & $S_{\alpha}$ & .810~\cite{gao2023thorough} & .833 & \blue{\textbf{0.1\%$\uparrow$}}\\
    \midrule
    \multirow{3}{*}{DUT-LF~\cite{duft}} & $F_\beta^{mean}$ & .906~\cite{gao2023thorough} & .936 & \red{\textbf{3.3\%$\uparrow$}} \\
                            & $E_\beta^{mean}$ & .954\cite{gao2023thorough} & .957 &  \red{\textbf{0.3\%$\uparrow$}}\\
                            & $S_{\alpha}$ & .911\cite{gao2023thorough} & .938 &  \red{\textbf{3.0\%$\uparrow$}} \\
    \hline
  \end{tabular}\\
  \caption{Quantitative comparison with other methods on different datasets in terms of $F_\beta^{mean}$, $E_\beta^{mean}$, and $S_{\alpha}$. We conduct an unequal comparison by selecting the highest scores from previous works, \ie, ``PreV'' and comparing them with our results.}
  \label{FES}
  \vspace{-2em}
\end{wraptable}
 \label{Measure}

The proposed method significantly surpasses this integrated benchmark. The network's performance is most effectively proved, particularly with the large-scale and richly varied PKU dataset. By establishing the pre-network connectivity and the in-network connectivity of LF data, the network reconnects the intrinsic relationships between different light field camera images, achieving a $23\%$ improvement in MAE compared with STSA$_3$. \revised{\textbf{It should be noted that the training dataset of STSA$_3$ is an extension dataset (DUT-LF $+$ HFUT $+$ PKU-LF). We used the PKU-LF dataset, and the network performance still exceeded by 23\%.} In Table \ref{FES}, we perform a comparison in terms of other evaluation criteria following PKU team~\cite{gao2023thorough}. While other networks may perform well in certain respects, LF Tracy still surpasses previous methods on a majority of metrics. This fully demonstrates the network's superior comprehensive perception capabilities without being data-dependent.}

\subsection{Qualitative Results} \label{4.3}
\begin{figure}[ht]
\vspace{-1.5em}
  \centering
    \begin{minipage}{.42\textwidth}
    \centering
    \includegraphics[width=0.96\linewidth]{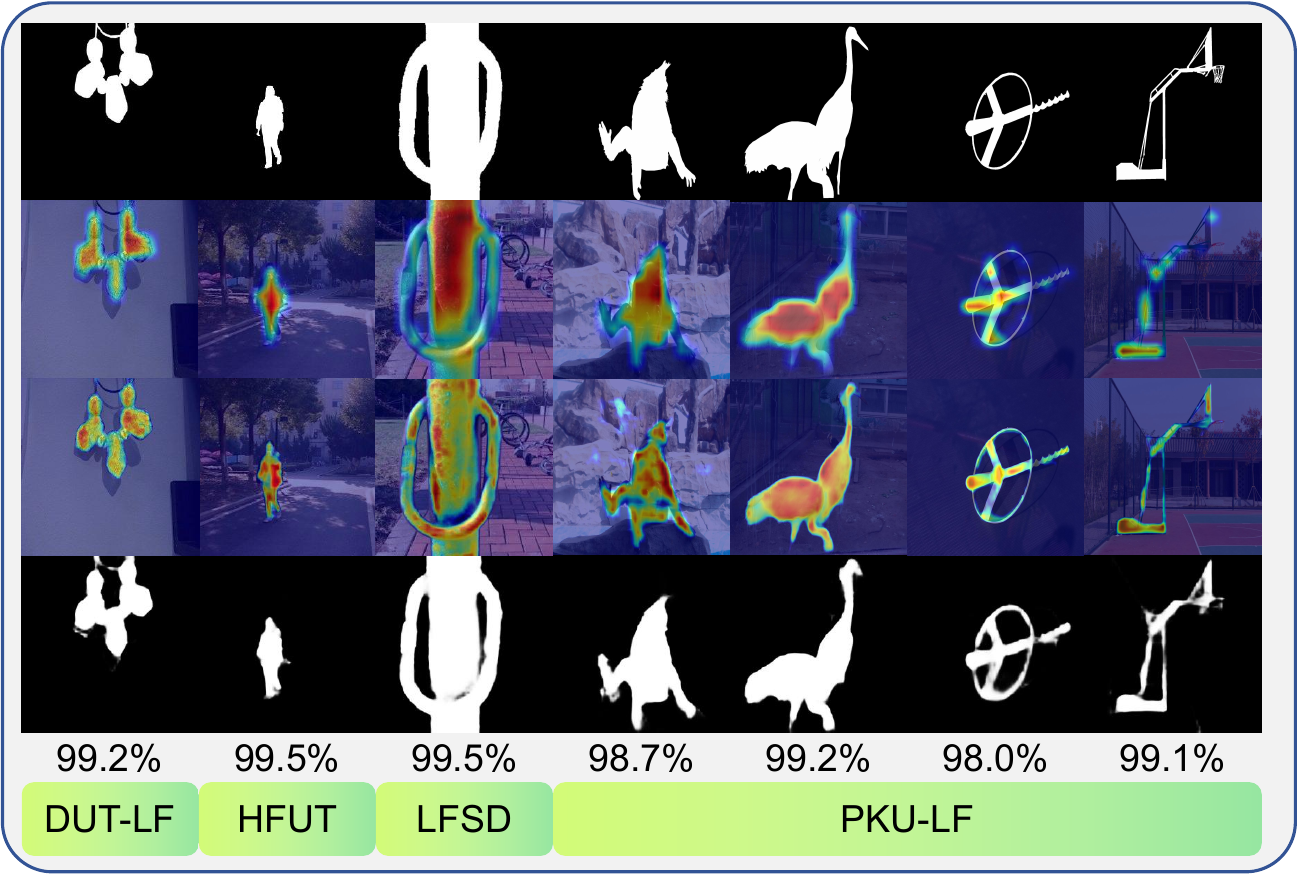}
    \caption{Qualitative Result on four datasets. From top to bottom, the ground truth, AF image feature maps, decoder maps, and predicted masks are illustrated.}
    \label{fig:Qualitative Result}
  \end{minipage}
  \begin{minipage}{.56\textwidth}
    \centering
    \includegraphics[width=0.96\linewidth]{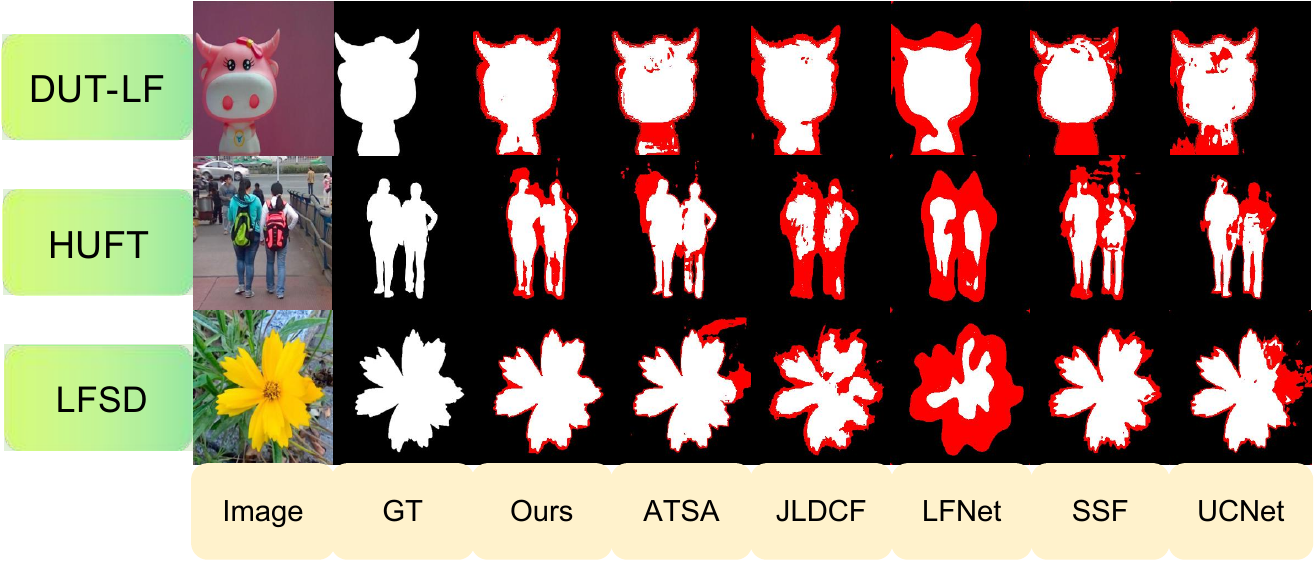} 
    \caption{Qualitative Comparison on three datasets. The difference maps between the visual results of various methods and the ground truth are displayed. Red pixels indicate pixels where the predicted results do not align with the ground truth.}
    \label{fig:Qualitative Comparision}
  \end{minipage}\hfill

\vspace{-1.5em}
\end{figure}

It can be seen from Fig.~\ref{fig:Qualitative Result} that the LF Tracy achieves outstanding accuracy across different scenarios by establishing intra-network and inter-network connectivity. Whether dealing with a single scene or complex scenarios, the network delivers excellent visualization results. Especially, for transparent backboards under varying lighting conditions, the network identifies the object through efficient information processing. Meanwhile, thin structures have always been a challenging issue in SOD tasks, yet the network has successfully identified both the necks of animals and the slender support poles of basketball hoops. Furthermore, the visual comparison results demonstrate the method's superiority, as in Fig.~\ref{fig:Qualitative Comparision}. The proposed network accurately identifies the locations of objects, and notably, it precisely identifies challenging boundaries and lines. For the images in the middle row, the area with two pedestrians walking side by side is particularly challenging to discern. The varied colors and textures of their clothing present a significant challenge to the network. While other methods show numerous errors in this region, the proposed network achieves accurate identification.

 \

\vspace{-3em}
\section{Ablation Studies}
\vspace{-0.5em}
In this section, \revised{several ablation studies are conducted to showcase the process of designing the network from scratch. Firstly, in Sec.~\ref{AB:1}, the experiments are carried out to comprehensively examine the effects of various components incorporated in the methods. Sec.~\ref{AB:3} showcases an in-depth analysis for the IA Module and FS Stack. Sec.~\ref{AB:4} investigates the performance of different backbones for the SOD task. Sec.~\ref{AB:2} demonstrated the in-depth analysis for MixLD. }
\vspace{-1em}

\vspace{-1.5em}
\subsection{Ablation Study for the Approach} \label{AB:1}

In the experimental analysis, as shown in Table.~{\ref{tab:51}}, we ablated components of the approach to assess their contributions. The optimal performance achieved an MAE of $0.046$. Firstly, eliminating the data augmentation strategy MixLD resulted in a performance decrease, and adapting CutMib~\cite{xiao2023cutmib} has few contributions to the performance. This indicates the necessity of MixLD to connect the different data before sending them into the network. After that, we ablate the core component of the network, the IA module, and the multi-scale features are directly fused. The MAE dramatically increased. 
The observed significant disparity of $0.286$ highlights the effectiveness of the IA module. This module is integral for effectively realigning and managing the data imbalance across diverse sources. In particular, it is pivotal in reducing data mismatching between LF and AF images, facilitating more effective data integration, and improving accuracy with one stream encoder. Finally, without FS, the result is further reduced.

\begin{table}
\vspace{-1em}
    \centering
    \renewcommand{\arraystretch}{1.5}
        \begin{tabular}{c:c:c:c:c:c}
        \toprule[0.5mm]
        \textbf{Model} & \textbf{Our} & \textbf{~w/o. MixLD} & \textbf{~w. CutMib} & \textbf{~w/o. IA} & \textbf{~w/o. LF} \\ 
        \midrule
        \textbf{MAE} & \textcolor{red}{.046} & .052 & .051 & .332 & .057 \\
        \bottomrule
        \end{tabular}
        \caption{The ablation study for LF Tracy.}
        \label{tab:51}
\vspace{-3em}
\end{table}

\textbf{Parameters Analysis:} The total Parameters of the designed LF Tray are $30M$. After removing the first stage in the IA module, the parameters decrease to $27M$. Furthermore, removing the entire IA module, the parameters fall into $24M$. With only 6$M$ parameters, the network is capable of intra-network data connections and efficient feature fusion. \revised{\textbf{GFlops and FPS:} When processing 12 FS images, \ie, handling a total of 13 light field images in a single training flow, the GFLOPs and FPS are 104.13 and 4.28, respectively. Without the IA module, these values are 84.8 GFLOPs and 4.73 FPS.}

\vspace{-1em}
\subsection{In-depth analysis for the IA Module and FS Stack} \label{AB:3}
\vspace{-0.5em}

To demonstrate the contribution of the FS stack and the alignment and fusion capabilities of the IA module for asymmetric data, the ablation studies are conducted from three different aspects. 

\begin{wraptable}{r}{0.48\textwidth}
\vspace{-2em}
    \centering
    \renewcommand{\arraystretch}{1.5}
        \setlength{\tabcolsep}{1.5mm}
        \begin{tabular}{c:c:c:c:c}
        \toprule[0.5mm]
        \textbf{Stack Size} & \textbf{2} & \textbf{3} & \textbf{5} & \textbf{12}\\ 
        \midrule
        \textbf{w/o. IA} & .137 & .141 & .205 & .332 \\
        \textbf{w. IA} & .051 & .051 & .049 & \red{.046} \\
        \bottomrule
        \end{tabular}
        \caption{An ablation study for the IA Module is conducted to evaluate its capabilities in terms of feature fusion and alignment.}
        \label{tab:521}
        \vspace{-2em}
\end{wraptable}

\revised{\textbf{\ding{192} Focal Stack Images:} We compared the discrimination ability of the network with and without the IA module, using $2$, $3$, $5$, and $12$ FS images, respectively.~As indicated in Table.~\ref{tab:521}, without the IA module, continuously stacking FS images does not enhance the network's capability; rather, it negatively impacts the network. With the addition of the IA module, the focused range and implicit angular information in the FS are utilized, increasing the network's discrimination ability.} 

\revised{\textbf{\ding{193} Fusion strategy in IA module:} In Table \ref{tab:522}, four different fusion strategies are compared. Firstly, we introduced an attention-based feature interaction process, accompanied by a set of learnable parameters, to achieve the fusion of information from different data sources. Then, we replaced this process with deformable cross attention~\cite{zhu2020deformable}. Subsequently, we directly add the features point by point. Finally, we utilized cross-attention for feature interaction, directly adding the interacted feature maps. Although the point-by-point addition method has achieved significant results in semantic segmentation tasks, it does not work effectively for SOD tasks. Likewise, the method of deformable cross attention also did not surpass the method we proposed.} 

\begin{table}[h]
\vspace{-1em}
    \centering
    \begin{minipage}{0.48\textwidth}
        \centering
        \begin{tabular}{c:c:c:c:c}
        \toprule[0.5mm]
        \textbf{Strategy} & \textbf{A\&PD} & \textbf{DA} & \textbf{ADD} & \textbf{A\&D}\\ 
        \midrule
        \textbf{MAE} & \textcolor{red}{.046} & .056 & {.332} & \textcolor{blue}{.048} \\
        \bottomrule
        \end{tabular}
        \caption{The exploration of different fusion strategies: A\&PD represents attention and dot product (with learnable parameters), DA represents deformable cross attention, ADD represents addition, and A\&D represents attention and addition.}
        \label{tab:522}
    \end{minipage}\hfill
    \begin{minipage}{0.48\textwidth}
        \centering
        \setlength{\tabcolsep}{1.5mm}
        \begin{tabular}{c:c:c:c:c}
        \toprule[0.5mm]
        \textbf{R\&Rate} & \textbf{1} & \textbf{$1/4$} & \textbf{$1/8$} & \textbf{$1/16$}\\ 
        \midrule
        \textbf{MAE} & .050 & .049 & .046 & .050 \\
        \bottomrule
        \end{tabular}
        \caption{In the first step of the IA Module, the reduction rate for the dimensions of Query and Key is evaluated. We perform channel compression at different scales. The MAE and GFlops are reported. `R\&Rate' indicates Reduction Rate.}
        \label{tab:523}
    \end{minipage}
\vspace{-2.7em}
\end{table}

\revised{\textbf{\ding{194} Reduction rate:} Last but not least, a set of experiments are conducted to deeply access the better hyper-parameters within IA module. Inspired by~\cite{han2023agent}, the dimensions of the Query and Key are compressed in the IA module. Four different reduction rates are chosen. As shown in Table \ref{tab:523}, over-reducing or under-reducing the channel can lead to performance degradation. The best option is to reduce the query and key dimensions to $1/8$ of the original size.}

\vspace{-1em}
\subsection{Selection of Various Backbones} \label{AB:4}
\vspace{-0.5em}

\begin{wraptable}{r}{0.55\textwidth}
\vspace{-2em}
    \centering
    \begin{tabular}{c:c:c:c:c}
    \toprule[0.5mm]
    \textbf{Backbone} & \textbf{B0} & \textbf{B1} & \textbf{B2} & \textbf{B4}\\ 
    \midrule
    \textbf{PVTv2~\cite{wang2022pvt}} & .120 & .097 & \red{.072} & .087 \\
    \textbf{AgentPVT~\cite{han2023agent}} & .153 & .137 & .142 & .145\\
    \bottomrule
    \end{tabular}
    \caption{\revised{An ablation study for the selection of the encoder backbone is conducted. B0, B1, B2, B4 indicate the backbone scales.}}
    \label{tab:531}
    \vspace{-2em}
\end{wraptable}

We conducted a series of experiments based on traditional datasets to assess the optimal feature extraction backbone. The PVTv2~\cite{wang2022pvt} and the agent attention~\cite{han2023agent} are selected. To prevent pre-trained weights from causing an unfair comparison in the selection of backbones, we conducted experiments for $100$ epochs without pre-trained weights. Table~\ref{tab:531} shows that the agent attention is ineffective for the dataset, and the performance on the LFSOD dataset does not improve with the increase in the number of parameters. Due to this reason, we have chosen PVTv2 as the backbone.

\vspace{-1em}
\subsection{In-depth analysis for MixLD} \label{AB:2}
\vspace{-0.5em}

\textbf{Interaction probability between texture and depth information:} \revised{In determining the optimal combination for incorporating depth information into AF images (FS2AF) and augmenting each FS image with texture information (AF2FS). Several experiments are designed with occurrence probabilities set at 0.1, 0.5, and 0.9. From the Table. \ref{tab:541}, it can be seen that: \ding{192} Assigning low occurrence probabilities (0.1) to both FS2AF and AF2FS minimally impacts the experimental outcomes, yet the performance metrics are analogous to those achieved with the CutMix augmentation technique. \ding{193} Excessive integration of depth information into AF images (probability set at 0.9 for FS2AF) leads to a significant loss of spatial information, affecting the network’s performance. \ding{194} While injecting spatial information into FS images improves the network’s ability to discriminate, excessive fusion can damage the valuable depth cues.}
\vspace{-1em}

\begin{table}[ht]
\begin{minipage}[t]{0.45\linewidth}
  \centering
  \begin{tabular}{c:c:c:c:cl}
    \toprule[0.5mm]
    \multicolumn{2}{c:}{\multirow{2}{*}{Selection Rate}} & \multicolumn{3}{c}{FS2AF} &  \\  \cmidrule(lr){3-5}
    \multicolumn{2}{c}{}                                & 0.1       & 0.5      & 0.9      &  \\
    \midrule
    \multirow{3}{*}{\begin{tabular}[c]{@{}c@{}}AF2FS\end{tabular}} & 0.1 & 0.51      & 0.55     & 0.57     &  \\

    & 0.5 & \red{0.46}      & 0.49     & 0.54     &  \\

    & 0.9 & 0.49      & 0.52     & 0.59     &  \\
    \bottomrule
  \end{tabular}
  \footnotesize
  \\[1em]
  \caption{Exploration for the occurrence probabilities of FS2AF and AF2FS.} 
  \label{tab:541}
\end{minipage}%
\hspace{0.5cm} % Adjust the space between the tables
\begin{minipage}[t]{0.5\linewidth}
  \centering
  \renewcommand{\arraystretch}{1.5}
  \setlength{\tabcolsep}{1mm}
  \begin{tabular}{c:c:c:c:c:c}
    \toprule[0.5mm]
    \textbf{$\alpha=\beta$} & \textbf{0.1} & \textbf{0.3} & \textbf{0.5} & \textbf{0.7} & \textbf{0.9} \\ 
    \midrule
    \textbf{MAE} & .053 & .049 & \red{.046} & .073 & .142 \\
    \bottomrule
  \end{tabular}
  \footnotesize
  \\[3em]
  \caption{Exploration of the blending rates in MixLD.} %
  \label{tab:542}
\end{minipage}
\vspace{-3em}
\end{table}

\textbf{Blending rate analysis: }To explore the optimal blending ratio of AF image and FS. We altered the parameter $\alpha$ in the first step, which involves blending one FS slice into AF images. Furthermore, in the second step, the parameter $\beta$ is adjusted to merge the blended AF image into FS. Due to the various combinations of $\alpha$ $\text{-}$ $\beta$ pair, we only experimented with a few combinations based on $\alpha{\text{=}}\beta$. 
As shown in Table~\ref{tab:542}, the optimal outcome is achieved with a blending rate of $0.5$. Notably, deviations from this ratio, either by increasing or decreasing the blending rate, result in a discernible decline in performance.

\vspace{-1em}
\section{Conclusion}
\vspace{-0.5em}
\textbf{Contribution:} In this paper, we present a unified single-stream method (LF Tracy) for salient object detection, bridging the inter-network and intra-network data connectivity. 
\textit{First}, we have designed an efficient IA module. 
This module effectively addresses the feature mismatching of different LF representations. In combination with a single-pipeline encoder, it enables intra-network data connectivity. Uniquely, our study tests the network's performance and achieves leading results on four distinct datasets.
\textit{Second}, we propose a data augmentation strategy for saliency object detection, specifically targeting inter-network connectivity. This method facilitates interaction among different channels of data, enhancing the network's discriminative ability. 

\textbf{Limitation and Further Work:} 
The task of salient object detection is sensitive to the choice of backbone, which sets it apart from other dense prediction tasks, such as semantic segmentation. Establishing a unified pixel-wise prediction framework is challenging and requires investigation in future work.

\textbf{Acknowledgment:} This work was supported in part by the National Natural Science Foundation of China (No. 62473139), in part by Helmholtz Association of German Research Centers, in part by the MWK through the Cooperative Graduate School Accessibility through AI-based Assistive Technology (KATE) under Grant BW6-03, and in part by Hangzhou SurImage Technology Co. Ltd.

\bibliographystyle{splncs04}
\bibliography{bib}

\end{document}